\documentclass[a4paper,twoside]{article}

\usepackage{cite}
\usepackage{amsmath,amssymb}
\usepackage{algorithmic}
\usepackage{graphicx}
\usepackage{CJKutf8}
\usepackage{textcomp}
\usepackage{xcolor}
\usepackage{times}
\usepackage{soul}
\usepackage{url}
\usepackage[utf8]{inputenc}
\usepackage[small]{caption}
\usepackage{multicol}
\usepackage{pslatex}
\usepackage{apalike}
\usepackage{algorithm2e}
\usepackage[bottom]{footmisc}
\usepackage{booktabs}
\usepackage[switch]{lineno}
\usepackage{tcolorbox}
\usepackage{multirow}
\newcommand{\eq}[1]{\begin{small}\begin{align}#1\end{align}\end{small}}
\usepackage{float}
\usepackage[greek,english]{babel}
\newcommand{\ci}{~\cite} \newcommand{\re}{~\ref} 

\DeclareMathOperator*{\argmax}{arg\,max}
\newtcolorbox{mymathbox}[1][]{colback=white, sharp corners, #1}
\usepackage{SCITEPRESS}

\begin{document}
\title{Efficient and Flexible Topic Modeling using Pretrained Embeddings and Bag of Sentences}
\author{\authorname{Johannes Schneider}
\affiliation{ University of Liechtenstein, Vaduz, Liechtenstein}
\email{johannes.schneider@uni.li}
}


\abstract{
    Pre-trained language models have led to a new state-of-the-art in many NLP tasks. However, for topic modeling, statistical generative models such as LDA are still prevalent, which do not easily allow incorporating contextual word vectors. They might yield topics that do not align well with human judgment. In this work, we propose a novel topic modeling and inference algorithm.  We suggest a bag of sentences (BoS) approach using sentences as the unit of analysis. We leverage pre-trained sentence embeddings by combining generative process models and clustering. We derive a fast inference algorithm based on expectation maximization, hard assignments, and an annealing process. The evaluation shows that our method yields state-of-the art results with relatively little computational demands. Our method is also more flexible compared to prior works leveraging word embeddings, since it provides the possibility to customize topic-document distributions using priors. Code and data is at \url{https://github.com/JohnTailor/BertSenClu}
}
\onecolumn \maketitle \normalsize \setcounter{footnote}{0} \vfill

\section{\uppercase{Introduction} }
Topic modeling is an actively research text-mining technique\ci{abd23,chu22} that analyzes a corpus to extract latent topics which describe a semantic concept. It dates back to the last millennium with early techniques such as PLSA\ci{hof99}, which compute word-topic and document-topic distributions. Back in 2003 Latent Dirichlet Allocation (LDA) \ci{ble03} marked a milestone in topic modeling. It is still the de-facto standard, i.e., many researchers and practitioners alike still rely on this technique or variants thereof for analyzing diverse text corpora, such as short \ci{hack20,sch21}, mid-sized \ci{han20}, and long texts\ci{sba20}. This is surprising for two reasons: (i) Topic models such as LDA fall short in many ways, and (ii) the field of natural language processing has rapidly advanced since the introduction of LDA in 2003. While LDA can be derived in a mathematically sound and elegant way from a few basic assumptions, many topics identified by LDA are meaningless and removed during manual investigation. Topics might change on a word by word basis resulting in unnatural topic assignments. One cause is the optimization objective of LDA, i.e., perplexity, which tends to assign high probability to frequent words not indicative of any particular topic such as `a',`the',`is',`are', etc. Perplexity is said to be a measure that does not agree as well with human judgment as other measures like PMI\ci{new10}. Thus, an issue of LDA is that the resulting topic models are not well aligned with human-defined topics. Furthermore, the bag of words(BoW) assumption of LDA combined with the generative modeling approach, leads to words within a single sentence being assigned to multiple different topics, e.g., topics can change after almost every word as also reported in \ci{gru07,sch18}. Neglecting word order is convenient from a mathematical point of view since it leads to independence of words given a topic. But it is inaccurate from a semantical perspective, e.g., it makes a big difference, if ``You eat a fish'' or ``A fish eats you''. While numerous works have attempted to address issues in LDA\ci{zha21s}, the overall success has been limited, especially considering the fact that natural language processing(NLP) has rapidly advanced since 2003. For once, (static) word vectors were popularized in 2013\ci{mik13} in NLP and later further developed into more dynamic, contextual vectors\ci{dev18}. These word vectors led to major improvements in many NLP tasks. They allow computing similarities between individual words or even documents in a fine-grained manner. This is an important step forward in NLP that priorly performed only binary comparisons, i.e., comparing if two words are identical or not. 
Despite the advantages of word vectors, for topic modeling only relatively few attempts have been undertaken to either include pre-trained static word vectors\ci{das15,mia16,ngu15}, contextualized vectors\ci{men22,gro22} or learn them from the corpus\ci{niu15,dien20}. If so, typically a BoW model\ci{gro22,mia16,ngu15,das15,niu15} was used often by extending LDA or building on variational inference\ci{mia16}. Our approach is to deviate from the classical BoW model, which is too fine-grained and leads to topic changes for potentially any word. We also refrain from an analysis of documents as a whole\ci{men22}, which neglects the possibility that words have multiple topics. We introduce a bag of sentence model (BoS) that splits the document into sequences of a few sentences. That is our elementary unit of analysis is a sequence of typically one to five sentences (rather than a single word), which is sufficient to identify a topic but not too large to contain multiple topics. A sequence is assigned to a single topic, i.e., we perform hard assignments rather than computing soft assignments, which has computational advantages. It is also simpler to understand for the end-user. In contrast to \ci{men22,gro22} we also forego any dimensionality reduction and show that this leads to favorable outcomes. Our inference algorithm leverages ideas from expectation maximization to compute topic-document probabilities. It can be understood with the help of the aspect model. To compute assignments of sentences to topics, we rely on extending K-Means from clustering a single set of data points to a set of sets, i.e., we use a cluster prior given by topic-document probabilities. In turn, our inference is orders of magnitude faster than complex deep learning models and more flexible for users than \ci{men22}. However, it is slower than \ci{gro22} which fails to discover multiple topics within a document and lacks in topic quality. Thus, in summary, our method yields state-of-the-art performance with modest computational costs, while being flexible to adjust to user needs.
Our contributions are as follows:\\
\begin{enumerate}
    \item We introduce a novel topic model and the bag of sentence model (BoS) that leverages pretrained sentence embeddings. It resembles a compromise between prior approaches relying on words, i.e., the bag of words, and documents as a whole, i.e., clustering of documents as a basis for topic modeling.
    \item We provide a novel inference mechanism using a form of ``annealing'' rooted in clustering and the aspect model. To extract topics, i.e., a ranking of words within a topic from sentence to topic assignments, we propose a simple and efficient measure combining frequency of a word within a topic and relative relevance of a word within a topic. This also leads to automatic exclusion of rare words and words not relevant for a topic.
    \item Through evaluation on multiple datasets and comparing against multiple methods, we show that our method enjoys modest computational demands while outperforming existing models based on multiple topic measures, i.e., topic coherence and topic coverage as measured using a downstream task. We also perform a holistic evaluation comparing other relevant aspects such as functionality of methods from an end-user perspective.
\end{enumerate}

\section{\uppercase{SenClu Model}}
Our model can be viewed from two persepctives. First, it can be seen as performing similarity-based clustering based on sentence embeddings and centroids being the mean of cluster points similar to k-Means++\ci{art07}. However, we also have cluster priors, i.e., topic-document probabilities. 

Second, our model can also be viewed from the aspect model\ci{hof01} which defines a joint probability of documents and words $D \times W$ to derive the topic-document probabilities. A key conceptual difference is that we deviate from the word frequency driven generative process (with its flaws) and that we treat (short sequences of) sentences as the smallest unit of analysis rather than single words. That is, a sequence of a few sentences is assigned to a topic. Furthermore, the probability of a sentence given a document is not the typical generative probability (as in LDA). Technically, it is a binary distribution, i.e., 0 or 1, that originates from cluster assignments. Like the standard aspect model we assume conditional independence of a sequence of sentences $g$ and a document $d$ given a topic $t$: 
	
\eq{	
		p(g,d):= p(d)\cdot p(g|d) \label{eq:dat}\\
		p(g|d):=\sum_t p(g|t)\cdot p(t|d) \label{eq:amod}	
}

A document $d$ is a sequence of sequences of sentences $d=(g_0,g_1,...,g_{n-1})$ with each $g_i=(s_{j},s_{j+1},...,s_{j+n_s})$ being a sequence of $n_s$ consecutive sentences and $j \in [i\cdot n_s,(i+1)\cdot n_s-1]$. Furthermore, $s_i$ is the $i$-th sentence in document $d$. In turn, each sentence $s_i=(w_0,w_1,...)$ is a sequence of words.\footnote{Typically, in topic modeling words also include numbers, but not any punctuation.}. The same word can occur multiple times in a sentence, i.e., it can happen that $w_i=w_j$ for $i\neq j$. Analogously, the same sentence can occur multiple times in a document.
The distribution $p(d)$ indicates the likelihood of a document. We do not estimate it, i.e., we simply use a uniform distribution as in other works, e.g.,\cite{sch18}. This is motivated by the assumption that all documents are considered roughly equally likely.  We define the remaining missing terms $p(g|t)$ and $p(t|d)$ of the aspect model (Equation \ref{eq:amod}) as follows:
	
\eq{
p(g|d)&=\frac{\max_{t} \{h(g,t)\cdot p(t|d)\}}{\sum_{t} \{h(g,t)\cdot p(t|d)\}} \propto\max_{t} \{h(g,t)\cdot p(t|d)\} \label{eq:m1}\\	
h(g,t) &:= cos(v_g,v_t) \label{eq:m1a}\\
p(t|d)& := \frac{\alpha+\sum_{i<|d|} 1_{t=\argmax_{t'}\{h(g_i,t')\cdot p(t'|d)\}}}{|d|\cdot(1+\alpha)} \label{eq:m2}
}
First, observe that both $p(g|d)$ and $p(t|d)$ are probability distributions. But they differ from the classical models (like LDA) that define the probabilities merely as a count based measure (of word occurrences).  
The term $p(g|d)$ is based on the similarity of the sequence and the topic vector $h(g,t)$ (Eq. \ref{eq:m1a}). 
To compute the probability of a sentence sequence in a document (Eq. \ref{eq:m1}), we also use $p(t|d)$ and the concept of latent variables, i.e., topics, as done in the aspect model (Equation \ref{eq:amod}). 

More concretely, we use the idea that a topic $t$ as well as a  (consecutive) sentences $g$ can be represented by a vector capturing semantic properties. We use a sentence vector $v_g$ from a pretrained sentence transformer, i.e., we use sentenceBERT\ci{rei19} in our evaluation and a topic vector $v_t$ computed based on the assigned sentence vectors $v_g$ to a topic. We average the sentence vectors $v_g$ assigned to topic $t$ to obtain the topic vector $v_g$.  The more similar the two vectors $v_g$ and $v_t$ are, the more related is topic $t$ to sentences in $g$. As similarity measure we use cosine similarity, which is common for high dimensional vectors, but the dot-product yields almost identical results. Equation \ref{eq:m2} states that the probability of a topic $t$ given a document is proportional to the number of sentence sequences $g_i$ assigned to $t$. We also added a smoothing constant $\alpha$ that gives a user some control over whether to prefer documents being assigned to few or many topics (similarly to the topic prior $\alpha$ in LDA). We discuss it in more depth later.




\section{\uppercase{Computing Word-topic Scores}}
The established norm for presenting topic model outcomes is listing the most likely words of the word-topic distribution $p(w|t)$. Since our method relies on sentences assignments, i.e., outputs $p(g|t)$, an analogous way would be to report the most likely sentences. However, this bears two disadvantages. First, users are not familiar with such outputs since it differs from all existing topic models. Second, sentences are lengthy, and some words in the sentence might be unrelated to the topic, possibly creating confusion.

Obtaining the empirical word-topic distribution $p(w|t)$ from $p(g|t)$ is straightforward since we assign each sentence to a topic and, in turn, each word to a topic. Thus, we can compute the relative frequency of a word within a topic. A shortcoming of this approach is that common words like ``is'',``are'',``a'',``the'', etc. obtain high probabilities within all topics since these words can occur in almost any sentence. Words occurring in all topics cannot be used to discriminate among topics. Words are more meaningful to a topic if most occurrences of a word are assigned to that topic only. Thus, we compute a score $score(w|t)$ for a word within a topic using a combination of frequency of a word within a topic and a measure capturing the relevance of the word to topic $t$ relative to other topics. The higher the score, the more relevant the word is for a topic. 

The frequency $n(w|t)$ is the number of words assigned to topic $t$, $n(w|d)$ the occurrences of $w$ in document $d$ and the frequency $n(w)$ of $w$ in the corpus is $n(w)=\sum_t n(w|t)$. The frequency score is computed based on a damped frequency, i.e., $\sqrt{\max(n(w|t)-n_{min},0))}$. Without damping, frequency tends to get too much importance.\footnote{Damping is common in NLP, e.g., taking the logarithm or square root of frequencies is also done in the computation of tf-idf}
The term $n_{min}$ can be seen as the frequency of words that originates by chance, i.e., from a uniform distribution, or from artificial distributions, e.g., a word occurs mostly just in one document. The term gives a minimum threshold based on the expectation of a word $n(w)/|T|$, if we assigned words uniformly at random to each topic, the standard deviation of $std(n(w|t))$ of $w$ across all topics $t$ and the maximum occurrences of $\max_d n(w,d)$ in a document. It is defined as $n_{min}:=n(w)/|T| + std(n(w|t))+\max_d n(w,d)$. That is, words should occur more than their expectation (for random assignments) plus the observed empirical standard deviation of assignments. If this does not hold, words are not indicative of a topic. Furthermore, the term $\max_d n(w,d)$ for the maximum occurrence of a word expresses the need that a word must occur in multiple documents. The idea is that topics should be prevalent in many documents rather than originating from one or very few documents.  
In addition to a frequency measure, we use a measure for ``relative'' relevance. It is the probability beyond expectation, i.e., $p(t|w)-1/|T|$. The probability $p(t|w)$ is approximated by the fraction of assignments of the word to the topic relative to its total occurrences. Thus, if a word occurs only in one topic, i.e., we have $p(t|w)=1$, then the word is highly relevant for the topic, even if it is a very rare word. If the word occurs with the same counts in every topic, we have $p(t|w)=1/|T|$. Such a word is not indicative of a topic and, therefore, it should have zero relevance. Indeed, the relative relevance is zero, i.e., $p(t|w)-1/|T|=0$. 
Overall, $score(w|t)$ is then the product of the frequency and relevance term, i.e., $\sqrt{n(w|t)-n_{min}}\cdot (p(t|w)-1/|T|)$. Words that have zero or negative scores are not reported. It can happen that a topic consists only of few words or even no word with positive scores. Commonly, this is the case if a cluster contains a few, very specialized words that might occur only in one or two documents, words that are fairly common across all topics (like "the","a") and a few topical words that occur only infrequently for the topic, i.e., there might be a sports cluster where the word sports occurs 1000 times and a cluster, where it occurs only say 5 times. In this case, both the relevance and frequency score are small. Formulas for all terms are stated also in Algorithm \ref{alg:SenClu}.

\begin{algorithm}[htp]
    \caption{SenClu(\small{docs $D$, nTopics $k$, Prior $\alpha$})}  \label{alg:SenClu}
    \begin{algorithmic}[1]
        \begin{small}
            \STATE $p(t|d):= 1/k$; $T:=[1,k]$; $epochs:=10$; $c(\alpha):=\max(8,\alpha)$, $n_s:=3$ 
            \STATE Tokenize docs $d \in D$ into a sequence of sentences $d:=(g_0,g_1,...)$. Each sentence group $g_i=(s_0,s_1,...,s_{n_s-1})$ is disjoint and consists of $n_s$ sentences. Each sentence $s_i$ is tokenized into words.
            \STATE Pretrained sentence embedder model $M:s\rightarrow v_s$ with dimension $d_s:=|v_s|$ \ci{rei19}
            \STATE Initialize topic vectors $v_t \in \{s | s \in d, d \in D\}$ using k-means++ and cosine similarity             
            \FOR{$i \in [1,epochs]$}            
            \STATE $A_{t,d}=\{\}$ \COMMENT{$\forall t \in T,d \in D$} \COMMENT{Begin of E-step}
            \FOR{$d \in D$}
            \STATE $r:=$ random number in $[0,1]$
            \STATE $i:= 1 \text{ if } r<0.5+i/(2\cdot epochs) \text{ else } 2$
            \FOR{$g \in d$}            
            \STATE $t_{g,d} = \argmax^i_{t} \{cos(v_g,v_t)\cdot p(t|d)\}$ \COMMENT{$\argmax^i$ gives the argument that yields the $i$-th largest value, i.e., 1 gives the largest}
            \STATE $p(t|g,d) =  1  \text{ if } t_{g,d}=t \text{ else } 0$
            \STATE $A_{t_{g,d},d}:=A_{t_{g,d},d} \cup g$            
            \ENDFOR
            \ENDFOR            
            \STATE $v_t:=\frac{\sum_{d \in D} \sum_{s \in A_{t,d}} v_s}{\sum_{d \in D} |A_{t,d}|}$ \COMMENT{Begin of M-step}
            \STATE $p(t|d):= \frac{|A_{t,d}|+c}{|d|+c}$
            \STATE $c(\alpha)=\max(c(\alpha)/2,\alpha)$            
            \ENDFOR                    
            \STATE \emph{Compute word-topic scores }
        \STATE $n(w|t)= \sum_{d \in D} \sum_{g \in A_{t,d}} \sum_{s \in g} \sum_{w_i \in s} 1_{w_i=w}$
        \STATE  $p(t|w):= \frac{n(w|t)}{\sum_{t} n(w|t)}$
        \STATE  $n_{min}:=n(w)/|T| + std(n(w|t))+\max_d n(w,d)$
        \STATE  $score(w|t):=\sqrt{n(w|t)-n_{min}}\cdot (p(t|w)-1/|T|)$        
        \end{small}
    \end{algorithmic}			
    
\end{algorithm}

\section{\uppercase{Inference}} \label{sec:inf}
We maximize the likelihood of the data  $\prod_d \prod_{g \in d} p(g,d)$ to estimate model parameters. To this end, we rely on our model definitions (Equations \ref{eq:dat}, \ref{eq:m1}, and \ref{eq:m2}). Generally, inference methods such as Gibbs sampling or variational inference are slow. Speeding up such methods, e.g., by integrating out (collapsing) variables of a Gibbs sampler, are difficult. We utilize expectation-maximization(EM) and clustering ideas to obtain a fast inference mechanism. In a similar fashion as \ci{sch18} we derive an EM algorithm based on standard probabilistic reasoning using sentence-topic assignment frequencies. EM performs two steps. The purpose of the E-step is to estimate latent variables, i.e., the probability $p(t|g,d)$ of a topic given sentence(s) $g$ in document $d$. In the M-step the topic distribution $p(t|g,d)$ is kept fixed and used to maximize the loss function with respect to the parameters. 
We perform hard assignments. That is, we assume that a sentence in a document has exactly one topic with probability one, and the probability of all other topics is zero. Thus, the topic $t_{g,d}$ of a sentence group $g$ in a document $d$ is the most probable topic: 

    \eq{		
    &t_{g,d} := \argmax_{t} \{h(g,t)\cdot p(t|d)\} \label{eq:ass} \\
    &p(t|g,d) = \begin{cases}  \label{eq:siass}
    1  \qquad t_{g,d}=t\\
    0 \qquad t_{g,d}\neq t\\
    \end{cases}	\\
    &A_{t',d}:=\{g|t_{g,d}=t', g \in d\}
    }

The term $A_{t,d}$ are the sentence groups $g$ assigned to topic $t$ in document $d$. Our definitions (Equation \ref{eq:ass} and following) differ from PLSA and LDA, where each word within a document is assigned a distribution typically with non-zero probabilities for all topics. This has computational advantages. But it also leads to challenges, e.g., getting stuck in local minima that require, for example, an annealing process using randomness to escape such minima that is continuously decreased.
In the M-Step, we want to optimize parameters. Analogously to Equations (9.30) and (9.31) in \ci{bis06} we define the function $Q(\Theta,\Theta^{old})$ for the complete data log likelihood depending on parameters  $\Theta$:
	
\eq{
&\Theta^{new} = \argmax_{\Theta} Q(\Theta,\Theta^{old}) \\ &\text{ with } 	Q(\Theta,\Theta^{old}):= \sum_{d,t}  p(t|D,\Theta^{old}) \log  p(D,t|\Theta)  \label{eq:opt}		
}
		
The optimization problem in Equation (\ref{eq:opt}) might be tackled using various methods, e.g., using Lagrange multipliers. Unfortunately, simple analytical solutions based on these approaches are intractable given the complexity of the model Equations (\ref{eq:m1} and following). However, one might also look at the inference of parameters $p(g|t)$ and $p(t|d)$  differently. Assume that we are given all sentences $A_{t,d}$ assigned to topic $t$ from document $d$ for each document $d \in D$ in a collection of documents $D$.  Then, we define the topic vector $v_t$ to be simply the mean: 
\eq{
v_t:=\frac{\sum_{d \in D} \sum_{g \in A_{t,d}} v_g}{\sum_{d \in D} |A_{t,d}|}
}

To obtain the other parameters, our frequentist inference approach uses the empirical distribution: The probability of a topic given a document equals the fraction of sentences of the document assigned to the topic. Under mild assumptions, the maximum likelihood distribution equals the empirical distribution (see, e.g., 9.2.2 in\ci{bar12}): 
\eq{
p(t|d)&:=\prod_{i<|d|} p(g_i|d) \text{ (Using Equation \ref{eq:m2})} \approx \frac{|A_{t,d}|}{|d|} 
}
We also add a value $c(\alpha):=c'\cdot\alpha$ for a user-defined value $\alpha$ and a value $c'$ that is decayed throughout optimization. Thus, ultimately, we use
\eq{
p(t|d)\approx \frac{|A_{t,d}|+c(\alpha)}{|d|+k\cdot c(\alpha)}
}

\begin{table*}[htp]
    \vspace{15pt} 
    \caption{Datasets, classes are human pre-defined categories. The Gutenberg dataset contains parts of books of multiple languages}\label{tab:ds}    
    \centering
    \begin{footnotesize}    
        \begin{tabular}{ l  c  c  c  c}
            \toprule
         Dataset & Docs & \#Words/doc & Vocabulary & Classes \\	
         \midrule
            New York Times (NYT) & 31,997 & 690 & 25,903 & 10\\ 
            20Newsgroups&18,625 & 122 &37,150 & 20\\ 
            Gutenberg& 9,417 & 1,768 & 290,215 & 50 \\
            Yelp & 29,820 &191 & 75,791& 1 \\
            \bottomrule
        \end{tabular}   
    \end{footnotesize}
\end{table*} 

\begin{table*}
		\caption{The 20 Newsgroups} \label{tab:top}
		\scriptsize
		\setlength\tabcolsep{3pt}
		\begin{tabular}{| p{\textwidth}  |}\hline 
			comp.graphics, comp.os.ms-windows.misc, comp.sys.ibm.pc.hardware, comp.sys.mac.hardware, comp.windows.x, rec.autos, rec.motorcycles, rec.sport.baseball, rec.sport.hockey, sci.crypt, sci.electronics, sci.med, sci.space, misc.forsale,talk.politics.misc,talk.politics.guns, talk.politics.mideast, talk.religion.misc, alt.atheism, soc.religion.christian \\ \hline
		\end{tabular}	
\end{table*}
The value $c(\alpha)$ serves a dual purpose. The first purpose is to improve the optimization process and avoid local minima and poor initialization. The value $c$ is non-fixed.  We start from a large value $c_0:=8$ and decay it in each epoch $i$ to a new value $c_i:=\max(c_{i-1}/2,\alpha)$. Thus, the mininum value is given by a user-defined value $\alpha$ explained later. The decaying process ensures that early non-optimal topic-document distributions do not have too much influence. Technically, it ensures that in the early epochs, all topics remain feasible by ensuring a non-zero probability $p(t|d)$ for each topic. Otherwise, it can happen for small $p(t|d)$ and short documents $d$ that early in the optimization all or most sentences of a document $d$ are assigned to a not yet well-determined topic $t$ that will significantly evolve. Initially, we use a uniform distribution for $p(t|d)$. But if most or all sentences of a document are assigned to one topic (after the first epoch) then $p(t|d)$ gets close to 1 or even one and, in turn, $p(t'|d)$, of other topics $t'$ gets close to 0 or even 0. Thus, as expressed in Equation \ref{eq:ass} the assignment to the not yet well-formed topic $t$ is likely not changed in later epochs, and some topics might not be considered any more since their probability has become zero. Thus, conceptually, early in the optimization, we do not yet fully trust the resulting topic to document assignments since they are subject to change. In turn, we smoothen the resulting probability $p(t|d)$. One might also view the decaying process as a form of (simulated) annealing since it fosters changes of topic sentence assignments initially but slowly makes them less likely by making the distribution $p(t|d)$ more concentrated.
The second purpose of $c(\alpha)$, i.e., the user-defined value $\alpha$, is to control whether in the final assignment of sentences in a document should consist of a few or many topics. A large $\alpha$ leads to a smoother, more uniform topic-document distribution, while a value of $\alpha$ close to 0 leads to a concentrated distribution. From a user's perspective, this is similar to the hyperparameter $\alpha$ in LDA that serves as a prior and impacts the concentration of the topic-document distribution. Generally, for short texts few topics are desirable per document, while longer texts should have more. However, the choice is user and data-dependent. The interpretation of $\alpha$ is that ``hypothetically'' $\alpha$ sentence groups are assigned to each topic. Thus, $\alpha$ ranges between 0 and $\infty$. Typical, values are 0 to the average number of sentence groups of a document.
We initialize topic vectors $v_t$ with randomly chosen sentence group vectors $v_s$, very similiar to k-means++\footnote{Full details are in code with the link in the abstract}. One relevant aspect for our method (as well as for k-means) is that it can get stuck in a local minimum. A topic vector, i.e., a centroid, might be based on an outlier, e.g., if a sentence differs very much from all others and the topic vector is set to that sentence. This is highly undesirable, as it would be better to just assign the outlier to the nearest (large) cluster or remove it. We deal with the problem by randomly assigning sentence groups from one cluster to its nearest neighboring cluster. Of course, this leads to many incorrect assignments. Thus, we also perform annealling, i.e., initally we assign a document to the second most likely cluster with probability about 0.5 and then decrease the probability to 0 within half of the epochs. We must stop the random assignments significantly before the last epoch to ensure that there is also time for convergence.    
Our method called SenClu is summarized in Algorithm \ref{alg:SenClu}. It takes as input a corpus $D$, the number of desired topics $k$, and a prior $\alpha\geq 0$ given the preference for few or many topics per document. The initial value of $c(\alpha)$ has limited impacted as long as it is above 2. The group size for sentences could be altered by the user but we based on the rationale that usually a topic extends a few sentences. That is, any value from 1 to 5 is reasonable and also leads to comparable results. A larger value tends to smoothen the impact of rare words or single sentences that cannot be assigned to any topic without additional context. A very large value, i.e., if the value is about the average number of sentences in a document, leads to the situation that topics are treated as a single unit.

\begin{table}
\vspace{-5pt}
\setlength\tabcolsep{1pt}
		\caption{Quantitative comparison for number of topics $k$ for SenClu. }\label{tab:resk}
    \centering	
		\begin{tabular} {| c | l | l  | l | }
			\hline			
			Dataset & nTopics $k$ & NMI  & PMI  \\ \hline
\multirow{3}{*}{20News} & 25 & 0.46\tiny{\text{$\pm$}0.01} & 0.8\tiny{\text{$\pm$}0.02} \\
&50 & 0.47\tiny{\text{$\pm$}.003} & 0.79\tiny{\text{$\pm$}.037} \\ 
 & 100 & 0.47\tiny{\text{$\pm$}0.0} & 0.73\tiny{\text{$\pm$}0.02} \\\hline
\multirow{3}{*}{Guten} & 25 & 0.2\tiny{\text{$\pm$}0.01} & 0.83\tiny{\text{$\pm$}0.07} \\
&50 & 0.2\tiny{\text{$\pm$}.003} & 0.67\tiny{\text{$\pm$}.03} \\ 
 & 100 & 0.2\tiny{\text{$\pm$}0.01} & 0.75\tiny{\text{$\pm$}0.03} \\\hline
\multirow{3}{*}{NYT} & 25 & 0.29\tiny{\text{$\pm$}0.01} & 0.77\tiny{\text{$\pm$}0.03} \\
&50 & 0.28\tiny{\text{$\pm$}.021} & 0.78\tiny{\text{$\pm$}.025} \\ 
 & 100 & 0.28\tiny{\text{$\pm$}0.01} & 0.75\tiny{\text{$\pm$}0.05} \\\hline
\multirow{3}{*}{Yelp} & 25 & - & 0.65\tiny{\text{$\pm$}0.03} \\
&50 & - & 0.62\tiny{\text{$\pm$}.007}\\ 
 & 100 & - & 0.6\tiny{\text{$\pm$}0.01} \\\hline
		\end{tabular}  
\end{table}

\begin{table}
\vspace{-5pt}
		\caption{Quantitative comparison for number of sentences $n_s$ for SenClu.}\label{tab:resns}
    \centering	
		\begin{tabular} {| c | l | l  | l | }
			\hline			
			Dataset & \#Sen. $n_s$ & NMI  & PMI  \\ \hline
\multirow{3}{*}{20News} &  1 & 0.45\tiny{\text{$\pm$}0.02} & 0.76\tiny{\text{$\pm$}0.02} \\
&3 & 0.47\tiny{\text{$\pm$}.003} & 0.79\tiny{\text{$\pm$}.037} \\ 
 &  9 & 0.47\tiny{\text{$\pm$}0.0} & 0.79\tiny{\text{$\pm$}0.03} \\\hline
\multirow{3}{*}{Guten} &  1 & 0.19\tiny{\text{$\pm$}0.01} & 0.66\tiny{\text{$\pm$}0.07} \\
&3 & 0.2\tiny{\text{$\pm$}.003} & 0.67\tiny{\text{$\pm$}.03} \\ 
 &  9 & 0.21\tiny{\text{$\pm$}0.02} & 0.76\tiny{\text{$\pm$}0.04} \\\hline
\multirow{3}{*}{NYT} &  1 & 0.28\tiny{\text{$\pm$}0.02} & 0.75\tiny{\text{$\pm$}0.03} \\
&3 & 0.28\tiny{\text{$\pm$}.021} & 0.78\tiny{\text{$\pm$}.025} \\ 
 &  9 & 0.31\tiny{\text{$\pm$}0.03} & 0.79\tiny{\text{$\pm$}0.03} \\\hline
\multirow{3}{*}{Yelp} &  1 & - & 0.65\tiny{\text{$\pm$}0.04} \\
&3 & - & 0.62\tiny{\text{$\pm$}.007}\\ 
 &  9 & - & 0.65\tiny{\text{$\pm$}0.04} \\\hline
		\end{tabular}  
\end{table}

\begin{table}
\vspace{-5pt}
		\caption{Quantitative comparison for prior $\alpha$ for SenClu. }\label{tab:resalpha}
    \centering	
		\begin{tabular} {| c | l | l  | l | }
			\hline			
			Dataset & $\alpha$ & NMI  & PMI  \\ \hline
\multirow{3}{*}{20News} & 0.25 & 0.43\tiny{\text{$\pm$}0.01} & 0.69\tiny{\text{$\pm$}0.03} \\
&2 & 0.47\tiny{\text{$\pm$}.003} & 0.79\tiny{\text{$\pm$}.037} \\ 
 & 8 & 0.49\tiny{\text{$\pm$}0.01} & 0.81\tiny{\text{$\pm$}0.02} \\\hline
\multirow{3}{*}{Guten} & 0.25 & 0.16\tiny{\text{$\pm$}0.01} & 0.73\tiny{\text{$\pm$}0.05} \\
&2 & 0.2\tiny{\text{$\pm$}.003} & 0.67\tiny{\text{$\pm$}.03} \\ 
 & 8 & 0.24\tiny{\text{$\pm$}0.01} & 0.78\tiny{\text{$\pm$}0.03} \\\hline
\multirow{3}{*}{NYT} & 0.25 & 0.24\tiny{\text{$\pm$}0.0} & 0.67\tiny{\text{$\pm$}0.0} \\
&2 & 0.28\tiny{\text{$\pm$}.021} & 0.78\tiny{\text{$\pm$}.025} \\ 
 & 8 & 0.33\tiny{\text{$\pm$}0.02} & 0.82\tiny{\text{$\pm$}0.03} \\\hline
\multirow{3}{*}{Yelp} & 0.25 & - & 0.58\tiny{\text{$\pm$}0.02} \\
&2 & - & 0.62\tiny{\text{$\pm$}.007}\\ 
 & 8 & - & 0.66\tiny{\text{$\pm$}0.02} \\\hline
		\end{tabular}  
\end{table}

\begin{table}
\vspace{15pt}
		\caption{Quantitative comparison between methods. Times are in minutes.}\label{tab:res}
    \centering
	\begin{footnotesize}
		\setlength\tabcolsep{2pt}    
		\begin{tabular} {| c | l | l  | l |l| }
			\hline			
			Dataset & Method & NMI  & PMI & Time  \\ \hline
            \multirow{4}{*}{20News} & BerTopic & 0.27 \scriptsize{$\pm$.011} & 0.2 \scriptsize{$\pm$.003} & 0.81 \scriptsize{$\pm$.012}\\
 & LDA & 0.24 \scriptsize{$\pm$.007} & 0.35 \scriptsize{$\pm$.002} & 0.31 \scriptsize{$\pm$.003}\\
 & TopClus & 0.38 \scriptsize{$\pm$.012} & 0.39 \scriptsize{$\pm$.016} & $>150$\\
 & \textbf{SenClu} & 0.47\tiny{\text{$\pm$}.003} & 0.79\tiny{\text{$\pm$}.037} & 2.26\tiny{\text{$\pm$}.031}\\ \hline
\multirow{4}{*}{Guten} & BerTopic & 0.09 \scriptsize{$\pm$.0} & 0.44 \scriptsize{$\pm$.02} & 1.6 \scriptsize{$\pm$.122}\\
 & LDA & 0.25 \scriptsize{$\pm$.007} & 0.36 \scriptsize{$\pm$.022} & 0.83 \scriptsize{$\pm$.001}\\
 & TopClus & 0.24 \scriptsize{$\pm$.004} & 0.35 \scriptsize{$\pm$.014} & $>150$\\
 & \textbf{SenClu} & 0.2\tiny{\text{$\pm$}.003} & 0.67\tiny{\text{$\pm$}.03} & 5.62\tiny{\text{$\pm$}.37}\\ \hline
\multirow{4}{*}{NYT} & BerTopic & 0.07 \scriptsize{$\pm$.009} & 0.2 \scriptsize{$\pm$.002} & 2.91 \scriptsize{$\pm$.05}\\
 & LDA & 0.21 \scriptsize{$\pm$.014} & 0.36 \scriptsize{$\pm$.008} & 1.24 \scriptsize{$\pm$.015}\\
 & TopClus & 0.25 \scriptsize{$\pm$.021} & 0.42 \scriptsize{$\pm$.009} & $>150$\\
 & \textbf{SenClu} & 0.28\tiny{\text{$\pm$}.021} & 0.78\tiny{\text{$\pm$}.025} & 6.47\tiny{\text{$\pm$}.158}\\ \hline
\multirow{4}{*}{Yelp} & BerTopic & - & 0.15 \scriptsize{$\pm$.008} & 0.96 \scriptsize{$\pm$.115}\\
 & LDA & - & 0.32 \scriptsize{$\pm$.011} & 0.37 \scriptsize{$\pm$.006}\\
 & TopClus & - & 0.36 \scriptsize{$\pm$.008} & $>150$\\
 & \textbf{SenClu} & - & 0.62\tiny{\text{$\pm$}.007} & 2.91\tiny{\text{$\pm$}.045}\\ \hline
 
		\end{tabular}  
	\end{footnotesize}	
\end{table}

\section{\uppercase{Evaluation}}
We perform a qualitative and quantitative evaluation on four benchmark datasets, four methods, and three metrics. We also assess the impact of parameters such as the number of topics $n_t$, the number of sentences $n_s$ (per group), and the prior $\alpha$.\\ 
\noindent\textbf{Settings:} We ran our experiments on an Ubuntu 20.4 system running Python 3.9 and Pytorch 1.13 running on a server with 64 GB of RAM, 16 cores of an AMD Threadripper 2950X CPU, and an NVIDIA RTX TI 2080 GPU. If not specified differently, we used $k=50$ topics, $\alpha=2$, and the top 10 words of each topic. For each configuration we performed three runs and report average and standard deviation.\\
\noindent\textbf{Methods:} LDA\ci{ble03} implemented in Python's Gensim 4.3\ci{reh11} served as a baseline since it is the most well-established topic modeling technique. BERTopic\ci{gro22} and TopClus\ci{men22} both use fixed pre-trained contextualized embeddings and some form of clustering, as explained in detail in related work. They represent the state-of-the-art with similar methodology.\\

\begin{table*}[ht]
   \vspace{15pt}
		\caption{Top 7 words of topics by SenClu and TopClus for first 15 of 50 topics }\label{tab:qua} 
 \scriptsize
 \setlength{\tabcolsep}{0.5pt}
		\begin{tabular}{|l| l | l |}\hline
		To.	&20Newsgroups Dataset	& New York Times Dataset \\ \hline
  \multicolumn{3}{|c|}{Method: SenClu}\\ \hline
0 & \scriptsize{hepatitis, biopsy, cph, chronic, hypoglycemia, pituitary, persistent} & \scriptsize{banquette, sauce, rum, cucumber, entree, menu, patronize}\\
1 & \scriptsize{infringe, participle, amendment, verb, indulge, infringing, constitution} & \scriptsize{tyson, boxing, heavyweight, bout, evander, knockout, holyfield}\\
2 & \scriptsize{pirating, protection, copy, cracked, pirated, cracker, disassemble} & \scriptsize{emission, dioxide, carbon, environmentalist, environmental, logging, landfill}\\
3 & \scriptsize{scsi, ide, drive, controller, bus, modem, mhz} & \scriptsize{japan, japanese, tokyo, nippon, mitsubishi, nomura, takeshita}\\
4 & \scriptsize{gld} & \scriptsize{prosecutor, trial, jury, defendant, judge, lawyer, juror}\\
5 & \scriptsize{doctor, medication, pain, hernia, diet, migraine, crohn} & \scriptsize{drug, patient, cancer, doctor, disease, health, dr}\\
6 & \scriptsize{satan, angel, heaven, enoch, god, eternal, poem} & \scriptsize{detective, police, arrested, stabbed, murder, arrest, graner}\\
7 & \scriptsize{wheelie, bike, aerobraking, landing, ride, bdi, riding} & \scriptsize{mir, astronaut, shuttle, nasa, module, atlantis, spacecraft}\\
8 & \scriptsize{window, graphic, microsoft, cica, adobe, rendering, shading} & \scriptsize{germany, german, deutsche, ackermann, frankfurt, dresdner, daimler}\\
9 & \scriptsize{solvent, bakelite, phenolic, wax, drying, adhesive, soldering} & \scriptsize{rate, economist, index, nikkei, bond, inflation, economy}\\
10 & \scriptsize{ei, ax, mq, pl, max, lj, gk} & \scriptsize{bedroom, apartment, bath, building, square, developer, ft}\\
11 & \scriptsize{xterm, motif, widget, server, mit, sunos, window} & \scriptsize{kerry, bush, mccain, clinton, presidential, president, poll}\\
12 & \scriptsize{israel, israeli, arab, palestinian, lebanese, palestine, gaza} & \scriptsize{cloning, gene, chromosome, genetic, cloned}\\
13 & \scriptsize{antenna, frequency, transmitter, radio, receiver, detector, khz} & \scriptsize{ounce, bullion, dollar, cent, mercantile, settled, crude}\\
14 & \scriptsize{airmail, mcwilliams, mcelwaine, dublin, expiration, dftsrv, albert} & \scriptsize{editor, circulation, magazine, reader, tabloid, publishing, journalism}\\ \hline
 
\multicolumn{3}{|c|}{Method: TopClus} \\ \hline
0 & \scriptsize{please, thanks, thank, appreciate, sorry, appreciated, gladly} & \scriptsize{student, educator, grader, pupil, teenager, adolescent, school}\\
1 & \scriptsize{saint, biblical, messiah, missionary, apostle, church, evangelist} & \scriptsize{surname, mustache, syllable, corps, sob, nickname, forehead}\\
2 & \scriptsize{iranian, korean, hut, child, algeria, vegetable, lebanese} & \scriptsize{participation, involvement, effectiveness, supremacy, prowess, responsibility}\\
3 & \scriptsize{considerable, tremendous, immense, plenty, countless, immensely, various} & \scriptsize{garage, dwelling, viaduct, hotel, residence, bungalow, building}\\
4 & \scriptsize{expression, phrase, symbol, terminology, prefix, meaning, coordinate} & \scriptsize{clit, lough, bros, kunst, mcc, quay, lund}\\
5 & \scriptsize{memoir, publication, hardcover, encyclopedia, bibliography, paperback} & \scriptsize{moth, taxa, una, imp, null, def, une}\\
6 & \scriptsize{anyone, somebody, anybody, someone, anything, everybody, something} & \scriptsize{many, everybody, anything, everyone, several, much, dozen}\\
7 & \scriptsize{individual, people, populace, human, being, inhabitant, peer} & \scriptsize{mister, iraqi, hussein, iraq, iranian, iran, kurdish}\\
8 & \scriptsize{disturbance, difficulty, complication, danger, annoyance, susceptible, problem} & \scriptsize{iraqi, iraq, baghdad, saddam, hussein, kuwait, iran}\\
9 & \scriptsize{beforehand, time, sooner, moment, waist, farther, halfway} & \scriptsize{dilemma, uncertainty, agitation, reality, dissatisfaction, implication, disagre.}\\
10 & \scriptsize{upgrade, availability, replacement, sale, modification, repository, compatibility} & \scriptsize{nominate, terminate, establish, stimulate, locate, replace, protect}\\
11 & \scriptsize{buy, get, install, spend, sell, keep, build} & \scriptsize{withstand, hesitate, imagine, explain, apologize, happen, translate}\\
12 & \scriptsize{appropriated, reverted, wore, abolished, rescued, exercised, poured} & \scriptsize{forefront, accordance, extent, instance, way, precedence, behalf}\\
13 & \scriptsize{government, diplomat, fbi, ceo, parliament, officer, parliamentary} & \scriptsize{privy, continual, outstretched, purposely, systematically, unused, unfinished}\\
14 & \scriptsize{graduation, university, rural, upstairs, overseas, basement, undergraduate} & \scriptsize{cautious, goofy, arrogant, painful, cocky, hasty, risky}\\
 \hline
		\end{tabular}
		\vspace{-6pt}
	\end{table*}

\noindent\textbf{Datasets and Pre-processing:}
For SenClu we tokenize documents into sentences using a simple rule-based tokenizer\ci{sad20}. The sentences are then transformed to contextual sentences embeddings using sentence transformers\ci{rei19}.
For LDA and to tokenize sentences, we used the gensim default tokenizer, while for BERTopic and TopClus pre-processing is included (as part of the library/repo). As post-processing for all methods we lemmatized topical words and removed duplicates before extracting the top 10 words. The datasets in Table\re{tab:ds} have already been used in the context of topic modeling \ci{sch18,men22} except for the Gutenberg dataset, which consists of books from different categories and languages from the public Gutenberg library (see upload). NYT comes with two human categorizations into 10 classes. We used locations, i.e., 10 pre-defined countries.


\subsection{Quantitative Evaluation}
We focused topic coherence and topic coverage. Coherent topics are more sensical. For topic coherence, we compute the normalized PMI score\ci{new10} on a document level as defined in \ci{sch18} using the English Wikipedia dump from 2022/10/01 as an external source. PMI score aligns better with human judgment than measures such as perplexity\ci{new10}.
For topic coverage, we utilize a downstream task, i.e., clustering based on topic models compared against predefined human categories as in \ci{men22} given us the normalized mutual information score(NMI).
We also report computation time since energy consumption is important in times of climate crises, and topic models are used by a diverse set of researchers and practitioners, where computation time matters. 
Computation time includes both training and inference of topics for a corpus. But not computing evaluation metrics for a given set of topics such as PMI, since it is the same for all methods. We include topic specific data pre-processing such as tokenization and computing embeddings, and the running time of the topic modeling, i.e., all steps needed to obtain word-topic scores and topic-document distributions.

\noindent \textbf{Results} 
Results for the sensitivity analysis of the (hyper)paraemeters are shown in Tables \ref{tab:resk}, \ref{tab:resns}, and \ref{tab:resalpha}. These parameters have some though limited impact, indicating that at least on a quantitative level, the algorithm behaves fairly insensitive to the parameters.\\
Results for comparing methods are shown in Table \ref{tab:res}. 
The quantitative comparison shows that in terms of topic outcomes, SenClu and TopClus are best. The fact that we outperform on topic coverage on two out of three datasets (Yelp lacks clustering information), i.e. a clustering task, against TopClus is surprising, since the method was explicitly optimized for clustering and evaluated on the same clustering task. We see this is as evidence that optimizing for clustering paired with dimensionality reduction is not needed for high quality topic modeling. On the contrary, it might hemper performance, since fundamental assumptions that would justify dimensionality reduction such as that embeddings are fairly noisy are not satisfied. Our method clearly outperforms on PMI computation, meaning that topic coherence is much larger. This can also be seen when looking at the actual topic words: Other models often have topics with common, non-topical words. A striking disadvantage of TopClus are the long computation times of several hours already for modest sized corpora. Our method SenClu is much faster but still requires considerable time, i.e., a few minutes, compared to LDA and BerTopic that often complete within about a minute. However, quantitative metrics are not sufficient to holistically assess the compared models as shown in the overall evaluation in Section \ref{sec:over} and the following qualitative evaluation.


\subsection{Qualitative Evaluation}
Table \ref{tab:qua} shows top words for the first 15 topics comparing against the method that performed best according to our evaluation and prior work\ci{men22}. We did not attempt to label topics, since during actual topic modeling users face exactly such outputs. However, for the 20Newsgroups dataset we listed the ground truth classes \ref{tab:top} to support understanding of topics and the dataset. Our assignment shows that TopClus somewhat suffers from the same issue as LDA: Common words might form topics that have little meaning and must be eliminated. For example, topic 0, 3, and 6 of the 20Newsgroups dataset consist of frequent non-topical words, while, for instance, topic 4,8 and 9 are not easy to assign to any topic. This is a common issue (as also observed for LDA)  nd rooted in the Bag of Words assumption. Other topics are well interpretable, e.g., topic 1 can be easily associated with the ground truth label `religion' (see Table  \ref{tab:top} for ground truth labels) and topic 11 with `forsale'. For SenClu most topics are easy to interpret, e.g., Topic 1 and 5 discuss a medical topic, Topic 3 hardware, etc. But it also contains a few topics, which make limited sense. For example, Topic 4 consists of just 1 token and topic 10 can also not well be interpreted. 

 \begin{table*}[htp]
    \vspace{15pt} 
    \caption{Summarized Assessment of Methods}\label{tab:meth}  
    \centering
        \begin{tabular}{ l  c  c  c  c c}
            \toprule
         Method & Multiple topics  & Topics per doc & Speed & Topic quality & Method complexity \\	
          &  per doc? &  controllable?  & & \\	
         \midrule
            LDA      & Y & Y & Fast & Medium & Low \\
            BerTopic & N & N & Fast & Medium-High & Low\\
            TopClus & Y  & N & Very Slow & High & Medium\\
            \textbf{SenClu}(ours) & Y & Y & Medium & High & Low\\                    
         \bottomrule
        \end{tabular}
   \vspace{-6pt}
\end{table*} 
\subsection{Overall Evaluation} \label{sec:over}
Table \ref{tab:meth} summarizes the comparison of all methods including quantitative evaluation but also offered functionality by the evaluated methods. Though LDA is very fast and a conceptually elegant approach, it suffers in terms of topic quality, which is the most important aspect of a topic model. Therefore, it is not the method of choice compared to methods relying on pretrained contextual embeddings. This finding is aligned with prior works \ci{men22,gro22}. BerTopic is very fast but topic quality is often not top notch and it treats documents as just having one topic. This is against the key idea of topic models that documents can have multiple topics. It is extremely problematic for long, diverse texts, where this is almost certainly the case.  TopClus yields high quality topics but suffers from challenges in interpretability, i.e., it uses a neural network with multiple loss functions and thousands of neurons. Neural networks are known to be challenging to interpret\ci{sch19c,mesk22,lon23}. More concerning is that its computational overhead makes TopClus tedious to apply in practice. Furthermore, users cannot express their intentions of how many topics are desirable per document. That is, if a user explicitly wants that documents should have a few topisc (or, on the contrary, many topics), the user cannot specify this. LDA and our method come with a hyperparameter ($\alpha$) guiding the algorithm to prefer few or many topics per document, which is a clear advantage. Thus, in summary, SenClu achieves state-of-the-art topic quality in reasonable time and it offers all functionality desirable by a user.

\section{\uppercase{Related Work} }\label{sec:rel}  
\noindent\textbf{Early, discrete topic models:} Already in the last millennium probabilistic latent semantic analysis (PLSA)\ci{hof99} was introduced as an improvement over latent semantic analysis using discrete word representations, i.e., one-hot word encodings. Latent dirichlect allocation (LDA) \ci{ble03} adds priors with hyperparameters to sample from distributions of topics and words as a generalization over LDA. LDA has been extended and varied in many ways. The generative probability of a word within a topic $p(w|t)$ is based on the relative number of occurrences of a word in a topic. In contrast, we use the idea that semantic similarity determines the probability $p(s|d)$ that a sentence, which is our unit of analysis rather than words as in LDA, belongs to a topic.
While most works rely on words, especially, a bag of words model, few models also investigated sentence assignments. For example, in \ci{gru07}, each sentence is assigned to a topic using a Markov chain to model topic-to-topic transitions after sentences. Though technically very different, the work also shares the idea that sentences are an important unit of analysis. Such an approach avoids that nearby words being assigned many different topics. \ci{bal16} is an extension of LDA that adds the extra ``plate'' of a sentence. All words of a sentence are assigned to the same topic. However, the model does not resolve the fundamental issue of LDA that frequency determines likelihood, e.g., short sentences of frequent words (and, in turn, frequent words) are most likely for a topic. \ci{sch18} used the idea to identify keywords that can influence the topic of nearby words, which effectively also leads to the situation that chunks of texts are assigned the same topic.\\
\noindent \textbf{Topic models with continuous word representations:} The aforementioned works treat words as one-hot encodings. With the success of (static) word vectors\ci{mik13} based on neural networks, the idea to use continuous representations also emerged in topic modeling with early approaches such as \ci{niu15,das15,mia16}. Although the idea to use external knowledge beyond the corpus to improve topic models has been known\ci{new11}, most works estimated vectors from the corpus used for topic modeling. Since then neural topic models that leverage deep learning networks for topic modeling have gained in popularity \ci{zha21s,dan22}. They have been used to address various issues such as accounting for correlated and structured topics\ci{xun17}, including meta-data and faster inference \ci{sri17}. \ci{bia20} specifically aimed to combine ProdLDA\ci{sri17} variant based on auto-encoders with document embeddings \ci{rei19}. The implementation is simple, i.e., adding document embeddings as part of the input to the autoencoder. The evaluation showed improvements compared to LDA\ci{ble03} and ProdLDA\ci{sri17}, but improvements were not consistent for other models. \ci{hoy20} employed knowledge distillation. They computed two distributions over words: One based on a standard topic model such as a variant of LDA\ci{sri17} and one based on a pretrained model, i.e., BERT. These two distributions serve as ground truth  for a student network that is trained to reconstruct documents. Most similar to our work are BERTopic\ci{gro22} and TopClus\ci{men22}. Both use pre-trained contextualized embeddings without any fine-tuning and some form of dimensionality reduction and clustering. Both works argue that dimensionality reduction is beneficial since distance measures tend to differ little when dimensions approach infinity \ci{gro22}. This is true, but it should be kept in mind that word embeddings\ci{rei19} only have about 300-800 dimensions and they were intentionally trained using these number of dimensions on large-scale data. Thus, in contrary, dimensionality reduction could lead to an information loss rather than be beneficial. \ci{men22} argues that training a dimensionality reduction layer optimized towards clustering leads to better outcomes since it facilitates clustering. Thus, in some sense, the burden of separating clusters is placed at least partially on the dimensionality reduction. Here, also dimensionality reduction might lead to an information loss. Furthermore, the proposed architecture for dimensionality reduction in \ci{men22} requires many parameters, which could lead to overfitting given that clustering corpora are often relatively small, e.g., existing of less than 100k documents. Thus, while dimensionality reduction is often a valuable technique in machine learning, we do not use dimensionality reduction of pre-trained embeddings due to the risk of information loss and the additional complexity.  In \ci{men22} a word is represented as the product of the pre-trained word embedding and an attention weight. Document embeddings are summed attention-weighted word embeddings. An objective is to optimally reconstruct these document sums by summing the topic embeddings of a document. The probability of a word belonging to a topic is given by a soft assignment using cosine similarity of the word embedding and the topic vector.  We differ by performing hard assignments, it enforces that a sentence is only assigned to one topic, which we deem more natural from a human topic modeling perspective. It also lessens the computational burden. In \ci{men22}, it was also realized that soft assignments might not be ideal. This issue dealt with squaring of the topic-word distribution, which amplifies the relative difference between the most likely and the second most likely topic. However, this is ad-hoc, since the power of 2 is arbitrary and constitutes a tuning parameter. The training in \ci{men22} requires three distinct losses, each loss must be weighed, which is more complex than our method conceptually and computationally more expensive. BERTopic\ci{gro22} simply computes embeddings of entire documents based on contextual word vectors\ci{rei19} and then clusters them using a density-based clustering technique, i.e., HDBScan. Thus, they lack the idea that a document can have multiple topics. Our and other models account for this option. In turn, our clustering is more similar to K-Means though as we also jointly compute topic-sentence probabilities, our clustering is more intricate. \\
\noindent\textbf{Pretrained language models:} First works on learning word vectors, a form of distributed representation, date back to the early 2000s \ci{ben00}. They were popularized about 10 years later using a simple neural network architecture that yielded static word vectors derived from large corpora, which allows doing arithmetics on words \ci{mik13}. Contextual word embeddings\ci{dev18} replaced static word vectors using much more intricate transformer architectures. They allowed deriving a vector based on contextual words, e.g. a word and nearby words in a text document. Since the first proposal, many possible improvements have been suggested, such as models tailored towards embedding sentences \ci{rei19},  robustness\ci{liu19}, and performance \ci{san19}. While using a sentence embedder is the obvious choice for a bag of sentence model, other models could also be used and could yield benefits, e.g., for faster inference words within a sentence might be aggregated using fast BERT version\ci{san19}.\\
\noindent\textbf{Topic labeling } discusses the problem of finding appropriate descriptions for identified topics or paragraphs. Typically, external knowledge such as Wikipedia is used to perform this task \ci{lau11}. Our approach uses only knowledge within the corpus as \ci{sch18,gro22} to identify topic words. On a high level, all follow a similar approach, i.e., weigh words based on a measure accounting for frequency of a term and distribution of a term (across topics) as done in classical measures such as term frequency and inverse document frequency (TF-IDF)\ci{has14}.


\section{\uppercase{Discussion and Future Work}}
Contextual word embeddings computed using transformers have pushed the state-of-the-art in NLP. However, their usage in topic modeling is still understudied. Our work introducing a bag of sentence model might also benefit from priorly derived insights for the bag of words, e.g., typically, nearby sentences are more related than more distance sentences. Leveraging such insights could improve the resulting topic models at the expense of a more complex algorithm. Complex and slow algorithms can be major obstacles for users since users might not have powerful machines but still desire to run the model with multiple hyperparameter settings. While our model is fast, in future work, we might further optimize its computation. For example, rather than using all documents to perform an EM step, we might rely on a subset. This could lead to faster convergence. Furthermore, our model could benefit from better sentence tokenizers and improvements in word embeddings.
We presented and derived our work from the perspective of the early but very foundational aspect model. An alternative approach is to view it from a clustering perspective, i.e., from k-Means. K-Means clustering also uses expectation maximization and computes cluster centers as the average of all points assigned to a cluster.  However, k-Means considers a set of points, whereas we consider sets of sets of points. This hierarchical representation influences the clustering process and requires extensions such as an ``annealing'' process to obtain high quality topics.

\section{\uppercase{Conclusions} }\label{sec:lim}
Topic modeling is challenging.  While many other NLP tasks have rapidly advanced in the last years, wildly used topic models still date back decades despite striking weaknesses. Our work has shown how to utilize external knowledge in the form of contextual word embeddings in an efficient manner to improve on prior topic models. It not only overcomes major shortcomings of prior works such as extremely long computation times or the inability to extract multiple topics per document, but it also improves on a variety of other measures such as downstream tasks. To do so, we utilize and introduce a novel topic model including inference mechanisms based on multiple ideas such as Bag-of-Sentences, hard assignments, and simulated annealing. Despite our promising experimental evaluation, we have elaborated in our discussion that  further improvements might be possible and encourage fellow researchers to engage in this challenge.

\bibliographystyle{apalike}
{\small
\bibliography{refs}}

\end{document}